\documentclass[10pt,twocolumn,letterpaper]{article}

\usepackage{cvpr}
\usepackage{times}
\usepackage{epsfig}
\usepackage{graphicx}
\usepackage{amsmath}
\usepackage{amssymb}
\usepackage{multirow}
\usepackage{booktabs}
\usepackage{caption}
\usepackage{tabularx}
\usepackage[T1]{fontenc}

\DeclareMathOperator*{\argmin}{arg\,min}

\usepackage[pagebackref=true,breaklinks=true,letterpaper=true,colorlinks,bookmarks=false]{hyperref}

\cvprfinalcopy %

\newcommand{\Best}[1]{\textbf{\textcolor{black}{#1}}}

\newcommand{\makecell}[2][@{}c@{}]{\begin{tabular}{#1}#2\end{tabular}}

\ifcvprfinal\fi
\begin{document}

\title{Model-blind Video Denoising Via Frame-to-frame Training}

\author{
\begin{tabularx}{\linewidth}{X}
\hfill \makecell{Thibaud Ehret} \hfill \makecell{Axel Davy} \hfill \makecell{Jean-Michel Morel} \hfill\null\\
\hfill \makecell{Gabriele Facciolo} \hfill \makecell{Pablo Arias} \hfill\null\\
\end{tabularx} \\
CMLA, ENS Cachan, CNRS\\
Universit\'e Paris-Saclay, 94235 Cachan, France\\
{\tt\small thibaud.ehret@ens-cachan.fr}
}

\twocolumn[{%
\renewcommand\twocolumn[1][]{#1}%
\maketitle
\begin{center}
\includegraphics[trim={15cm 6cm 5cm 2cm},clip,width=0.24\textwidth]{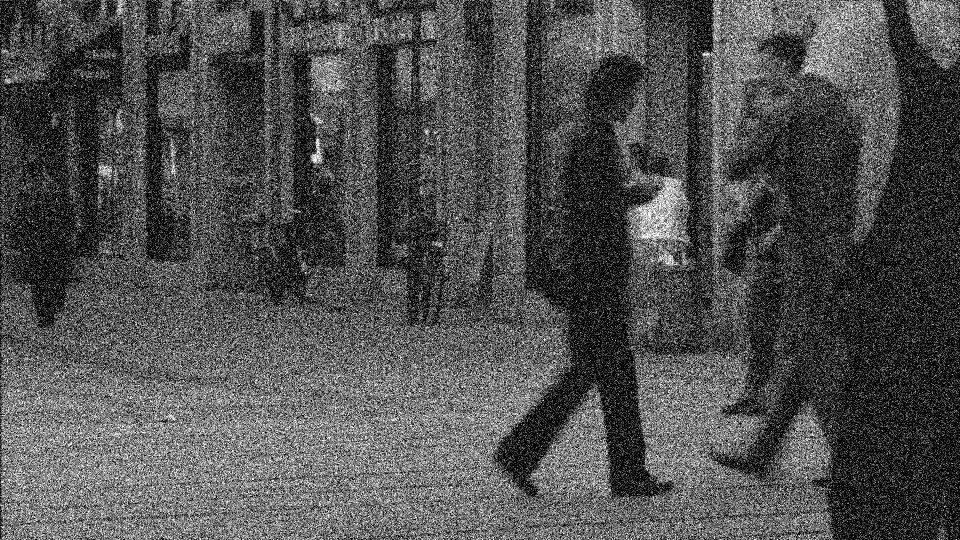}
\includegraphics[trim={15cm 6cm 5cm 2cm},clip,width=0.24\textwidth]{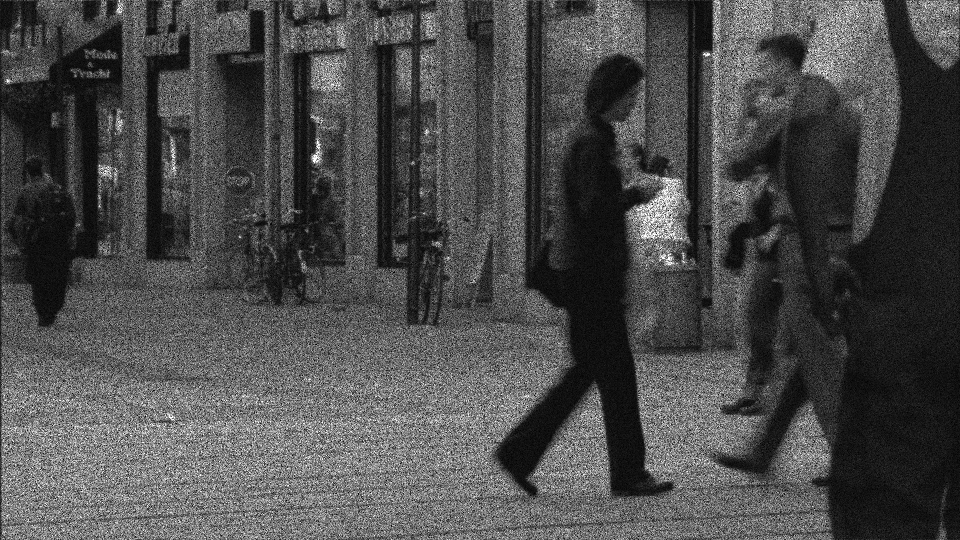}
\includegraphics[trim={15cm 6cm 5cm 2cm},clip,width=0.24\textwidth]{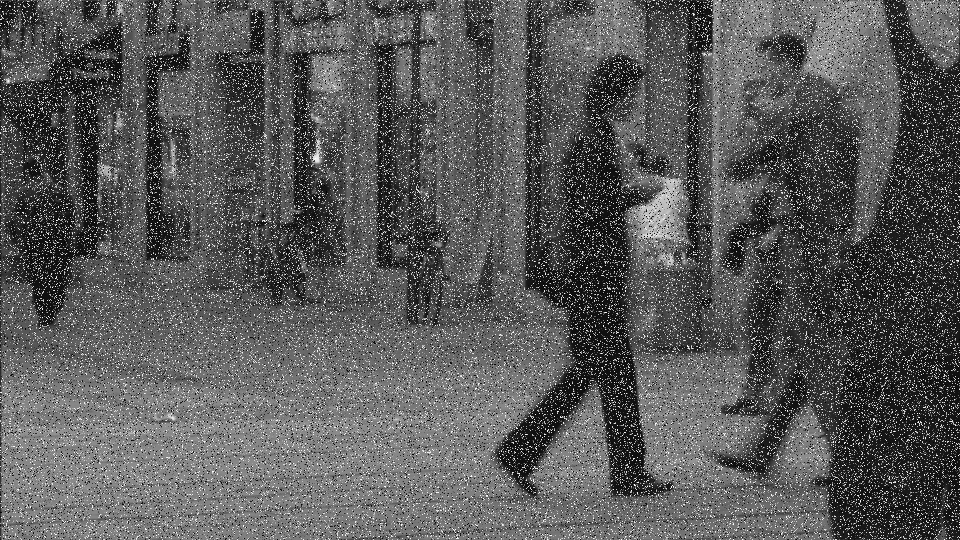}
\includegraphics[trim={15cm 6cm 5cm 2cm},clip,width=0.24\textwidth]{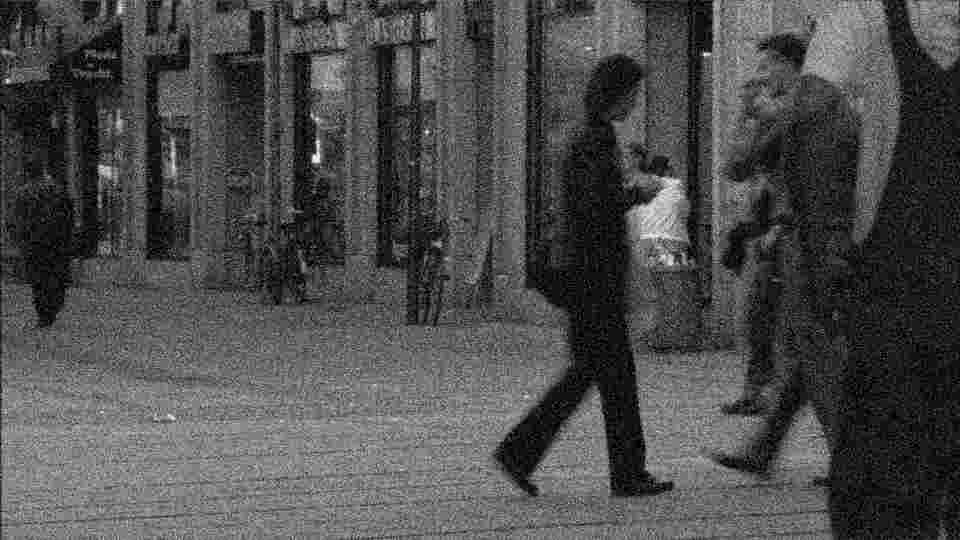}
\includegraphics[trim={15cm 6cm 5cm 2cm},clip,width=0.24\textwidth]{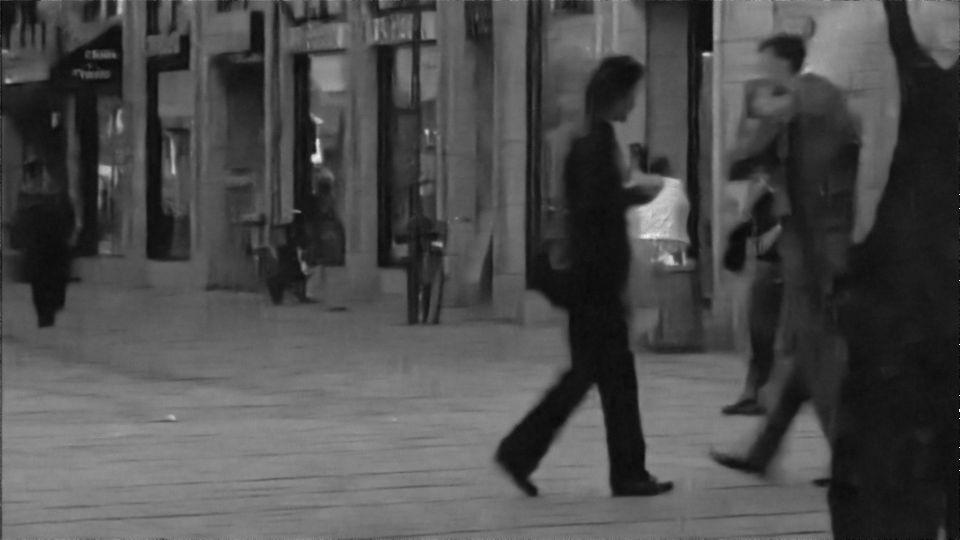}
\includegraphics[trim={15cm 6cm 5cm 2cm},clip,width=0.24\textwidth]{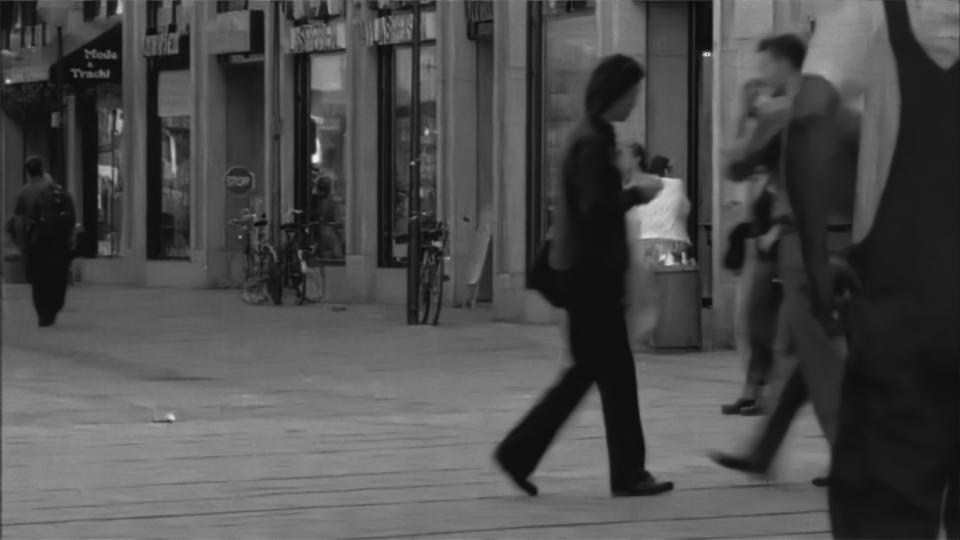}
\includegraphics[trim={15cm 6cm 5cm 2cm},clip,width=0.24\textwidth]{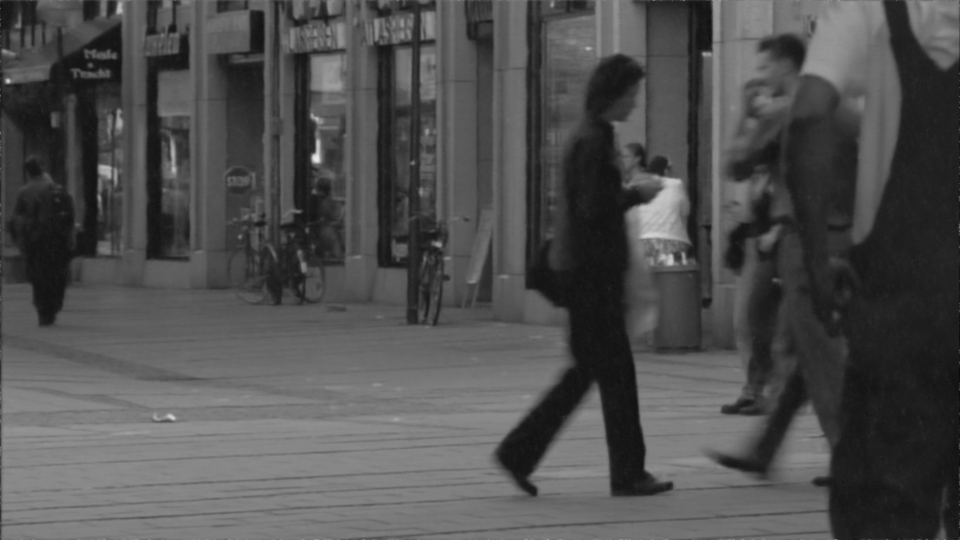}
\includegraphics[trim={15cm 6cm 5cm 2cm},clip,width=0.24\textwidth]{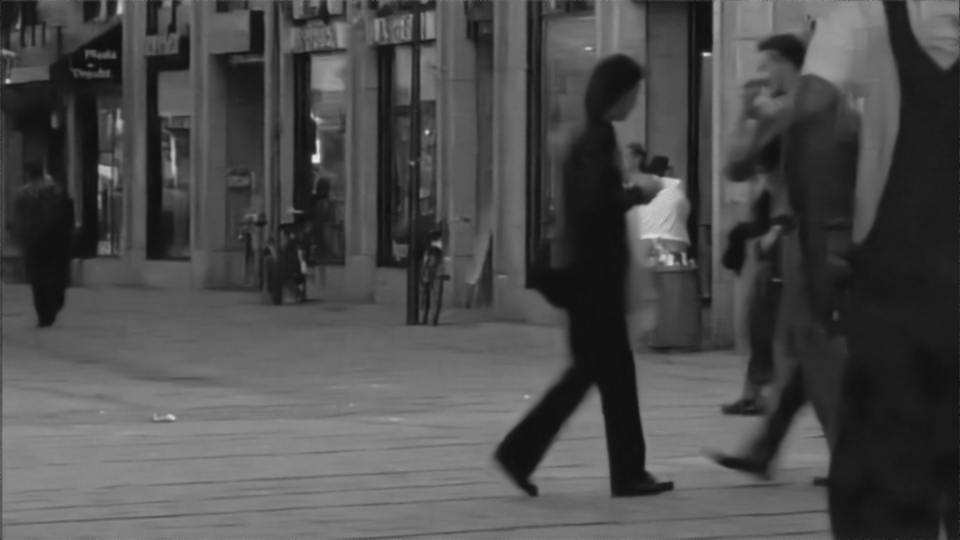}
\captionof{figure}{From the \emph{same} starting point and \emph{only} using the video, our fine-tuned network is able to denoise different noises without any artifact. The top images are the noisy and the bottom ones the denoised. From left to right: Gaussian noise, Poisson type noise, salt and pepper type noise and JPEG compressed Gaussian noise.}
\label{fig:teaser}
\end{center}
}]
\maketitle

\begin{abstract}
Modeling the processing chain that  has  produced a video is a difficult reverse engineering  task,  even when the camera is available. This makes model based video processing a still more complex task.  In this paper we propose a fully blind video denoising method, with two versions off-line and on-line. This is achieved by fine-tuning a pre-trained AWGN denoising network to the video with a novel frame-to-frame training strategy. Our denoiser can be used without knowledge of the origin of the video or burst and the post-processing steps applied from the camera sensor. The on-line process only requires a couple of frames before achieving visually pleasing results for a wide range of perturbations. It nonetheless reaches state-of-the-art performance for standard Gaussian noise, and  can be used off-line with still better performance.
\end{abstract}

\section{Introduction}

Denoising is a fundamental image and video processing problem. While the performance of denoising methods and imaging sensors has steadily improved over decades of research, new challenges have also appeared. 
High-end cameras still acquire noisy images in low lighting conditions. High-speed video cameras use short exposure times, reducing the SNR of the captured frames.
Cheaper, lower quality sensors are used extensively, for example in mobile phones or surveillance cameras, and require denoising even with a good scene illumination.

A plethora of approaches have been proposed for image and video denoising: PDE and variational methods \cite{rudin1992nonlinear,chambolle1997image}, bilateral filters \cite{tomasi1998bilateral}, domain transform methods \cite{Moulin1999,Portilla2003}, non-local patch-based methods \cite{buades2005non}. In the last decade, most research focused on modeling image patches \cite{Zoran2011,Yu2012,elad2006image} or groups of similar patches \cite{Dabov2006,mairal2009non,Lebrun2013,gu2014weighted,burger2012image}.
Recently the focus has shifted towards neural networks.

The first neural network with results competitive with patch-based methods was introduced in \cite{burger2012image}, and consisted of a fully connected network trained to denoise image patches. More recently, \cite{Zhang2017BeyondDenoising} proposed DnCNN a deep CNN with 17 to 20 convolutional layers with $3\times 3$ filters and reported a significant improvement over the state-of-the-art. The authors also trained a blind denoising network that can denoise an image with an unknown noise level $\sigma\in[0,55]$, and a multi-task network that can handle blindly three types of noise. A lighter version of DnCNN was proposed in \cite{17-zhang-ffdnet}, which allows a spatially variant noise variance by adding the noise variance map $\sigma^2(x)$ as an additional input. The architectures of DnCNN and FFDnet keep the image size throughout the network. Other networks have been proposed \cite{16-mao-red,16-santhanam-rbdn,18-chen-see-in-the-dark} that use pooling and up-convolutional layers in a U-shaped architecture \cite{Ronneberger2015U-Net:Segmentation}.
Other works proposed neural networks with an architecture obtained by unrolling optimization algorithms such as those used for MAP inference with MRFs probabilistic models \cite{09-barbu-arf,14-schmidt-csf,Chen2017TrainableRestoration,16-vemulapalli-deep-gcrf}.
For textures formed by repetitive patterns, non-local patch-based methods still perform better than ``local'' CNNs. To remedy this, some attempts have been made to include the non-local patch similarity in a CNN framework \cite{17-qiao-tnlrd,Chen2017TrainableRestoration,17-lefkimmiatis-nlcnn,18-yang-bm3dnet,18-cruz-nn3}.

The most widely adopted assumption in the literature is that of additive white Gaussian noise (AWGN). This is justified by the fact that the noise generated by the photon count process at the imaging sensor can be modeled as Poisson noise, which in turn can be approximated by AWGN after a variance stabilizing transform (VST) \cite{anscombe1948transformation,makitalo2011optimal,makitalo2011closed}.
However, in many practical applications the data available is not the raw data straight from the sensor. The camera output is the result of a processing pipeline, which can include quantization, demosaicking, gamma correction, compression, etc.  The noise at the end of the pipeline is spatially correlated and signal dependent, and it is difficult to model. Furthermore the details of the processes undergone by an image or video are usually unknown. To make things even more difficult, a large amount of images and videos are generated by mobile phone applications which apply their own processing of the data (for example compression, filters, or effects selected by the user). The specifics of this processing are unknown, and might change with different releases.

The literature addressing this case is much more limited. The works  \cite{lebrun2015noise,18-gonzales-denoising-decompression} address denoising noisy compressed images. RF3D \cite{maggioni2014joint} handles correlated noise in infrared videos. 
Data-driven approaches provide an interesting alternative when modeling is not challenging. CNNs have been applied successfully to denoise images with non-Gaussian noise \cite{Zhang2017BeyondDenoising,18-chen-see-in-the-dark,guo2018toward}.
In applications in which the noise type is unknown, one could use \emph{model-blind} networks such as DnCNN-3 \cite{Zhang2017BeyondDenoising} trained to denoise several types of noise, or the blind denoiser of \cite{guo2018toward}. These however have two important limitations. First, the performance of such \emph{model-blind} denoising networks very often drops with respect to \emph{model-specific} networks \cite{Zhang2017BeyondDenoising}.
Second, training the network requires a dataset of images corrupted with each type of noise that we wish to remove (or the ability to generate it synthetically \cite{guo2018toward}). Generating ground truth data for real photographs is not straightforward \cite{Plotz2017,18-chen-see-in-the-dark}. Furthermore, in many occasions we do not have access to the camera, and a single image or a video is all that we have.

In this work we show that, for certain kinds of noise, in the context of video denoising one video is enough: a network can be trained from a single noisy video by considering the video itself as a dataset. 
Our approach is inspired by two works: the one-shot object segmentation method \cite{Cae+17} and the noise-to-noise training proposed in the context of denoising by \cite{lehtinen2018noise2noise}. 

The aim of one-shot learning is to train a classifier network to classify a new class with only a very limited amount of labeled examples.
Recently Caelles \etal~\cite{Cae+17} suggested a one-shot framework for object segmentation in video, where an object is manually segmented on the first frame and the objective is to segment it in the rest of the frames. Their main contribution is the use of a pre-trained classification network, which is fine-tuned to a manual segmentation of the first frame. This fine-tuned network is then able to segment the object in the rest of the frames. This generalizes the one-shot principle from classification to other types of problems.
Borrowing the concept from \cite{Cae+17}, our work can be interpreted as a one-shot blind video denoising method: a network can denoise an unseen noise type by fine-tuning it to a single video. 
In our case, however, we do not require ``labels'' (i.e. the ground truth images without noise). Instead, we benefit from the noise-to-noise training proposed by \cite{lehtinen2018noise2noise}: a denoising network can be trained by penalizing the loss between the predicted output given a noisy and a second noisy version of the same image, with an independent realization of the noise.
We benefit from the temporal redundancy of videos and use the noise-to-noise training  between adjacent frames to fine-tune a pre-trained denoising network. That is, the network is trained by minimizing the error between the predicted frame and the past (or future) frame. The noise used to pre-train the network can be very different from the type of noise in the video.

We present the different tools, namely one of the state-of-the-art denoising network DnCNN \cite{Zhang2017BeyondDenoising} and a training principle for denoising called noise2noise \cite{lehtinen2018noise2noise}, necessary to derive our refined model in Section \ref{sec:tools}. We present our truly blind denoising principle in Section \ref{sec:blind}. We compare the quality of our blind denoiser to the state of the art in Section \ref{sec:experiments}. Finally we conclude and open new perspectives for this type of denoising in Section \ref{sec:conclusion}.

\section{Preliminaries}
\label{sec:tools}

The proposed model-blind denoiser builds upon DnCNN and the noise-to-noise training. In this section we provide a brief review of these works, plus some other related work.

\subsection{DnCNN}

DnCNN \cite{Zhang2017BeyondDenoising} was the first neural network to report a significant improvement over patch-based methods such as BM3D~\cite{Dabov2006} and WNNM~\cite{gu2014weighted}. It has a simple architecture {
inspired by the VGG network} \cite{Simonyan2014VeryRecognition}, consisting of 17 convolutional layers.
The first layer consists of 64 $3\times 3$ followed by ReLU activations and outputs $64$ feature maps. The next 15 layers also compute 64 $3\times3$ convolutions, followed by batch normalization \cite{ioffe2015batch} and ReLU. The output layer is simply a $3\times 3$ convolutional layer.

To improve training, in addition to the batch normalization layers, DnCNN uses \emph{residual learning}, which means that network is trained to predict the noise in the input image instead of the clean image. The intuition behind this is that if the mapping from the noisy input $f$ to the clean target $u$ is close to the identity function, then it is easier for the network to learn the \emph{residual mapping}, $f\mapsto f-u$. 

DnCNN provides state-of-the-art image denoising for Gaussian noise with a rather simple architecture. For this reason we will use it for all our experiments. 

\subsection{Noise-to-noise training}

The usual approach for training a neural network for denoising (or other image restoration problems) is to synthesize a degraded image $f_i$ from a clean one $u_i$ according to a noise model. Training is then achieved by minimizing the empirical risk which penalizes the loss between the network prediction $\mathcal F_{\theta} (f_i)$ and the clean target $u_i$. This method cannot be applied for many practical cases where the noise model is not known. In these settings, noise cannot be synthetically added to a clean image. One can generate noisy data by acquiring it (for example by taking pictures with a camera), but the corresponding clean targets are unknown, or are hard to acquire~\cite{chen2018image,Plotz2017}. 

Lehtinen \etal~\cite{lehtinen2018noise2noise} recently pointed out that for certain types of noise it is possible to train a denoising network from pairs of noisy images $(f_i, g_i)$ corresponding to the same clean underlying data and independent noise realizations, thus eliminating the need for clean data. This allows learning networks for noise that cannot be easily modeled (an appropriate choice of the loss is still necessary though so that the network converges to a good denoising).

Assume that the pairs $(f,u)$ are distributed according to $p(f,u) = p(u|f)p(f)$. For a dataset of infinite size, the empirical risk of an estimator $\mathcal F$ converges to the Bayesian risk, i.e. the expected loss: $\mathcal R(\mathcal F) = \mathbb E_{f,u}\{\ell(\mathcal F(u),f)\}$. The optimal estimator $F^*$ depends on the choice of the loss. From Bayesian estimation theory \cite{kay1993fundamentals} we know that:\footnote{The median and mode are taken element-wise. For a continuous random variable the $L_0$-loss is defined as a limit. See \cite{kay1993fundamentals} and \cite{lehtinen2018noise2noise}.}
\begin{align}
\ell = L_2 \quad \Rightarrow \quad & \mathcal F^*(f) = \mathbb E\{u\vert f\}\\
\ell = L_1 \quad \Rightarrow \quad & \mathcal F^*(f) = \text{median}\{u\vert f\}\\
\ell = L_0 \quad \Rightarrow \quad & \mathcal F^*(f) \approx \text{mode}\{u\vert f\}
\end{align}
Here $\mathbb E\{u|f\}$ denotes by the expectation of the posterior distribution $p(u|f)$ given the noisy observation $f$. During training, the network learns to approximate the mapping $f\mapsto F^*(f)$.

The key observation leading to noise-to-noise training is that the same optimal estimators apply when the loss is computed between $\mathcal F(f)$ and $g$, a second noisy version of $u$. In this case we obtain the mean, median and mode of the posterior $p(g\vert f)$. Then, for example if the noise is such that $\mathbb E\{g\vert f\} = \mathbb E\{u\vert f\}$, then the network can be trained by minimizing the MSE loss between $F(f)$ and a second noisy observation $g$. If the median (resp. the mode) is preserved by the noise, then the $L_1$ loss (resp. the $L_0$) loss can be used.

\section{Model-blind video denoising}
\label{sec:blind}

In this section we show how one can use a pre-trained denoising network learned for an arbitrary noise and fine-tune it to other target noise types using a single video sequence, attaining the same performance as a network trained specifically for the target noise for classic noise. This fine tuning can be done off-line (using the whole video as a dataset) or on-line, i.e. frame-by-frame, depending on the application and the computational resources at hand. 

As we will show in Section \ref{sec:experiments}, starting from a pre-trained network is key for the success of the proposed training, as we do not have a large dataset available as in \cite{lehtinen2018noise2noise}, but only a single video sequence. The use of a pre-trained network is in part motivated by works on transfer learning such as Zamir \etal~\cite{zamir2018taskonomy}. Denoising different noise models are related tasks. Our intuition is that a part of the network focuses on the noise type while the rest encodes features of natural images.

Our approach is inspired by the one-shot video object segmentation approach of \cite{Cae+17}, where a classification network is fine-tuned using the manually segmented first frame, and then applied to the other frames. As opposed to the segmentation problem, we do not assume that we have a ground truth (clean frames). Instead, we adapt the noise-to-noise training to a single video.

We need pairs of independent noisy observations of the same underlying clean image. 
For that we take advantage of the temporal redundancy in videos: we consider consecutive frames as observations of the same underlying clean signal transformed by the motion in the scene. To account for the motion we need to estimate it and warp one frame to the other.
We estimate the motion using an optical flow. We use the TV-L1 optical flow  \cite{zach2007duality} with an implementation available in \cite{sanchez2013tv}. This method is reasonably fast and is quite robust to noise when the flow is computed at a coarser scale.

Let us denote by $v_t$ the optical flow from frame $f_t$ to frame $f_{t-1}$. The warped $f_{t-1}$ is then $f^w_{t-1}(x) = f_{t-1}(x + v_t(x))$ (we use bilinear interpolation). Similarly, we define the warped clean frame $u^w_{t-1}$. We assume 
\begin{enumerate}
\item[\textit{(i)}] that the warped clean frame $u^w_{t-1}$ matches $u_t$, i.e. $u_t(x) \approx u^w_{t-1}(x)$, and 
\item[\textit{(ii)}] that the noise of consecutive frames is independent.
\end{enumerate}

Occluded pixels in the backward flow from $t$ to $t-1$ do not have a correspondence in frame $t-1$. Nevertheless, the optical flow assigns them a value. We use a simple occlusion detector to eliminate these false correspondences from our loss.
A simple way to detect occlusions is to determine regions where the divergence of the optical flow is large \cite{buades2016patch}.  We therefore define a binary occlusion mask as 
\begin{equation}
\kappa_t(x) = 
\begin{cases}
0 \text{ if } |\text{div}\, v_t(x)| > \tau \\ 
1 \text{ if } |\text{div}\, v_t(x)| \leq \tau.
\end{cases}
\label{eq:occ_mask}
\end{equation}
Pixels with an optical flow that points out of the image domain are considered occluded. In practice, we compute a more conservative occlusion mask by dilating the result of Eq. \eqref{eq:occ_mask}.

We then compute the loss masking out occluded pixels. For example, for the $L_1$ loss we have:
\begin{equation}
\label{eq:maskedl1loss}
\ell_{1}(f,g,\kappa) = \sum_x \kappa(x)\left|f(x) - g(x)\right|.
\end{equation}
Similarly one can define masked versions of other losses. In the noise-to-noise setting, the choice of the loss depends on the properties of the noise \cite{lehtinen2018noise2noise}. The noise types that can be handled by each loss in noise-to-noise have a precise characterization (the mean/median/mode of the \emph{noisy posterior} $p(g|f)$ have to be equal to those of the \emph{clean posterior} $p(u|f)$). Verifying this requires some knowledge about noise distribution.
In the absence of such knowledge, since the method is reasonably fast, an alternative would be to test different losses and see which one gives the best result.

In principle our method is able to deal with the same noise types as noise-to-noise. In practice we have some limitation imposed by the registration as it degrades for severe noise. For this reason we do not show examples with non-median preserving noise requiring the $L_0$ loss. For all our experiments we use the masked $L_1$ loss since it has better training properties than the $L_2$ \cite{Zhao2017LossNetworks}. Most relevant noise types often encountered in practice (poisson, jpeg-compressed, low-freq. noise) can be handled by the $L_1$ loss and the registration.

We now have pairs of images $(f_t, f^w_{t_1})$ and the corresponding occlusion masks $\kappa_t$ and we apply the noise-to-noise principle to fine-tune the network on this dataset. 
In order to increase the number of training samples the symmetric warping can also be done, i.e. warping $f_{t+1}$ to $f_t$ using the forward optical flow from $f_t$ to $f_{t+1}$. This allows to double the amount of data used for the fine-tuning.
We consider two settings: off-line and on-line training.

\paragraph{Off-line fine-tuning.} We denote the network as a parametrized function $\mathcal F_\theta$, where $\theta$ is the parameter vector. In the off-line setting we fine-tune the network parameters $\theta$ by doing a fixed number $N$ of steps of the minimization of the masked loss over all frames in the video:
\begin{equation}
\theta^{\text{ft}} = \argmin_\theta^{N,\theta_0} \sum_{t = 1}^T \ell_1(\mathcal F_\theta(f_{t}), f^w_{t-1}, \kappa_t)
\end{equation}
where by $\displaystyle\argmin_\theta^{N,\theta_0} E(\theta)$ we denote an operator which does $N$ optimization steps of function $E$ starting from $\theta_0$ and following a given optimization algorithm (for instance gradient descent, Adam~\cite{kingma2014adam}, etc.).
The initial condition for the optimization is the parameter vector of the pre-trained network. The fine-tuned network is then applied to the rest of the video.

\paragraph{On-line fine-tuning}
In the on-line setting we train the network in a frame-by-frame fashion. As a consequence we denoise each frame with a different parameter vector $\theta_t^{\text{ft}}$. At frame $t$ we compute $\theta_t^{\text{ft}}$ by doing $N$ optimization steps corresponding to the minimization of the loss between frames $t$ and $t-1$:
\begin{equation}
\theta_t^{\text{ft}} = \argmin_\theta^{N,\theta_{t-1}^{\text{ft}}} \ell_1(\mathcal F_\theta(f_{t}), f^w_{t-1}, \kappa_t).
\end{equation}
The initial condition for this iteration is given by the fine-tuned parameter vector at the previous frame $\theta^{\text{ft}}_t$.
The first frame is denoised using the pre-trained network. The fine-tuning starts for the second frame.
A reasonable concern is that the network overfits the given realization of the noise and the frame at each step. This is indeed the case if we use a large number of optimization iterations $N$ at a single frame. A similar behavior is reported in \cite{ulyanov2018deep}, which trains a network to minimize the loss on a single data point. We prevent this from happening by using a small number of iterations (e.g. $N=20$). We have observed that the parameters fine-tuned at $t$ can be applied to denoise any other frame without any significant drop in performance. 

The on-line fine-tuning addresses the problem of \emph{lifelong learning} \cite{zamir2018taskonomy} by continuously adapting the network to changes in the distribution of noise and signal. This is particularly useful when the statistics of the noise depend on time-varying parameters (such as imaging sensors affected by temperature).

\section{Experiments}
\label{sec:experiments}

In this section we demonstrate the flexibility of the proposed fine-tuning blind denoising approach with several experimental results. For all these experiments the starting point for the fine-tuning process is a DnCNN network trained for an additive white Gaussian noise of standard variation $\sigma=25$. In all cases we use the same hyper-parameters for the fine tuning: a learning rate of $5.10^{-5}$ and $N = 20$ iterations of the Adam optimizer. For the off-line case we use the entire video. The videos used in this section come from Derf's database\footnote{https://media.xiph.org/video/derf/}. They've been converted to grayscale by averaging the three color channels and downscaled by a factor two in each direction to ensure that they contain little to no noise. The code and data to reproduce the results presented in this section are available on \url{https://github.com/tehret/blind-denoising}.

To the best of our knowledge there is not any other blind video denoising method in the literature.
We will compare with state-of-the-art methods on different types of noise. 
Most methods have been crafted (or trained) for a specific noise model and often a specific noise level.
We will also compare with an image denoising method proposed by Lebrun \etal~\cite{lebrun2015noise} which assumes a Gaussian noise model with variance depending on the intensity and the local frequency of the image. This model was proposed for denoising of compressed noisy images. We cannot compare with some more recent blind denoising methods, such as \cite{chen2018image}, because there is no code available. We will compare with DnCNN~\cite{Zhang2017BeyondDenoising} and VBM3D~\cite{Dabov2007v}. VBM3D is a video denoising algorithm. All the other methods are image denoising applied frame-by-frame (perspectives for videos are mentioned in Section \ref{sec:conclusion}).

\begin{figure}
\centering
\includegraphics[width=\columnwidth]{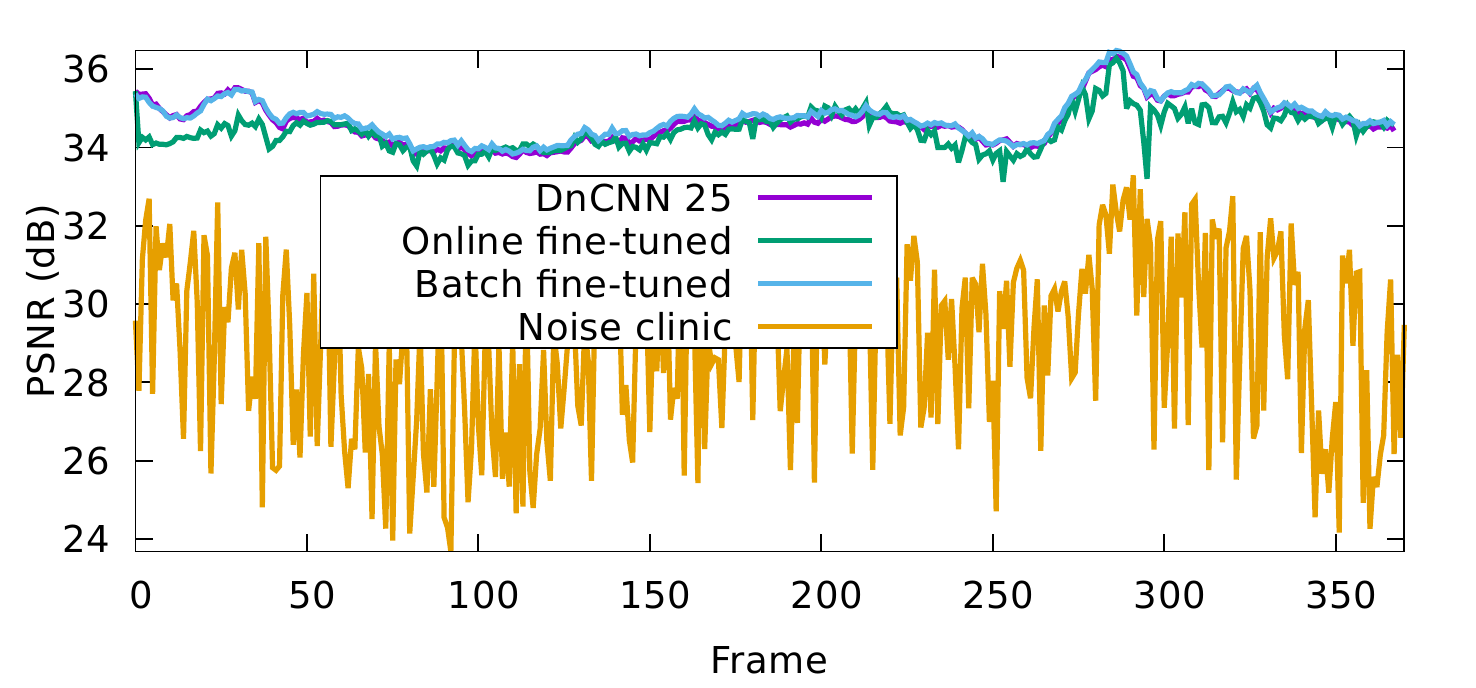}
\includegraphics[width=\columnwidth]{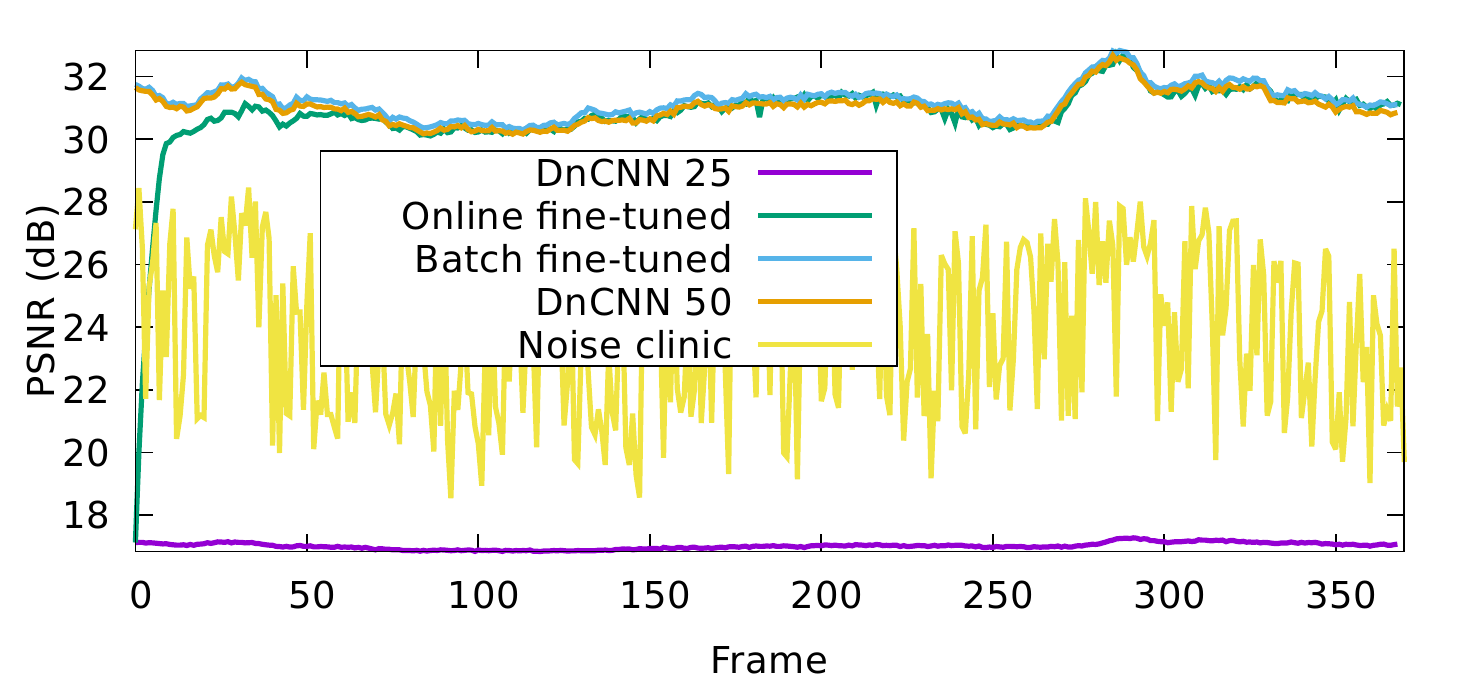}
\caption{The fine-tuning process is done on a sequence corrupted by an additive Gaussian noise of standard deviation $\sigma=25$ (top) or $\sigma=50$ (bottom). The fine-tuned networks (offline and online) achieve comparable performance than the reference networks.}
\label{fig:exp_gaussian}
\end{figure}

The goal of the first experiment is to compare against reference networks trained for these noises the regular way. The per-frame PSNRs are presented in Figure \ref{fig:exp_gaussian}. We applied the proposed learning process to a sequence contaminated with AWGN with standard deviation $\sigma=25$, which is precisely the type of noise the network was trained on and verified that it does not deteriorate the pre-training. The off-line fine-tuning performs on par with the pre-trained network. The PSNR of the on-line process has a higher variance, with some significant drops for some frames. 
For $\sigma=50$, we can see that both fine-tuned networks perform better than the pre-trained network for $\sigma=25$. In fact their performance is as good as the DnCNN network trained specifically for $\sigma=50$ (actually the off-line trained performs even slightly better than the reference network). Our process also outperforms the ``noise clinic'' of \cite{lebrun2015noise}.

\begin{figure*}
\centering
\includegraphics[width=\columnwidth]{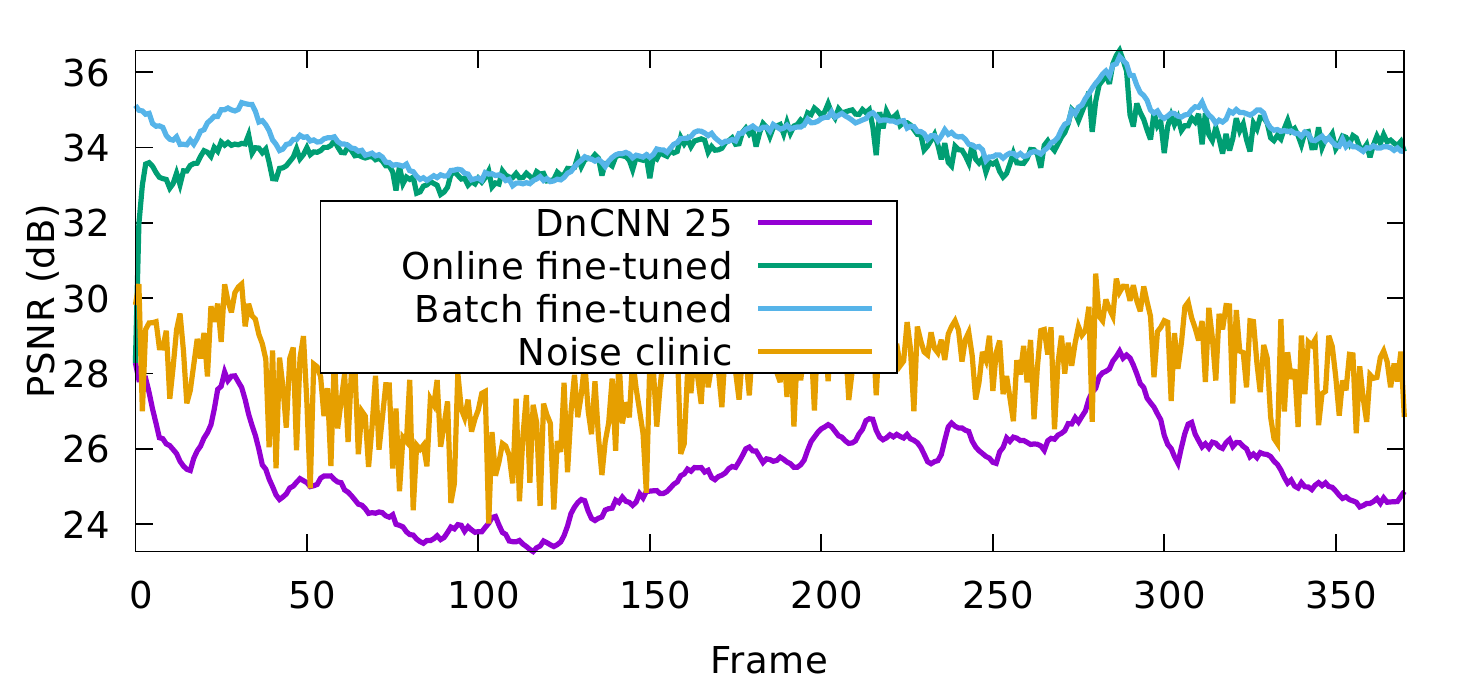}
\includegraphics[width=\columnwidth]{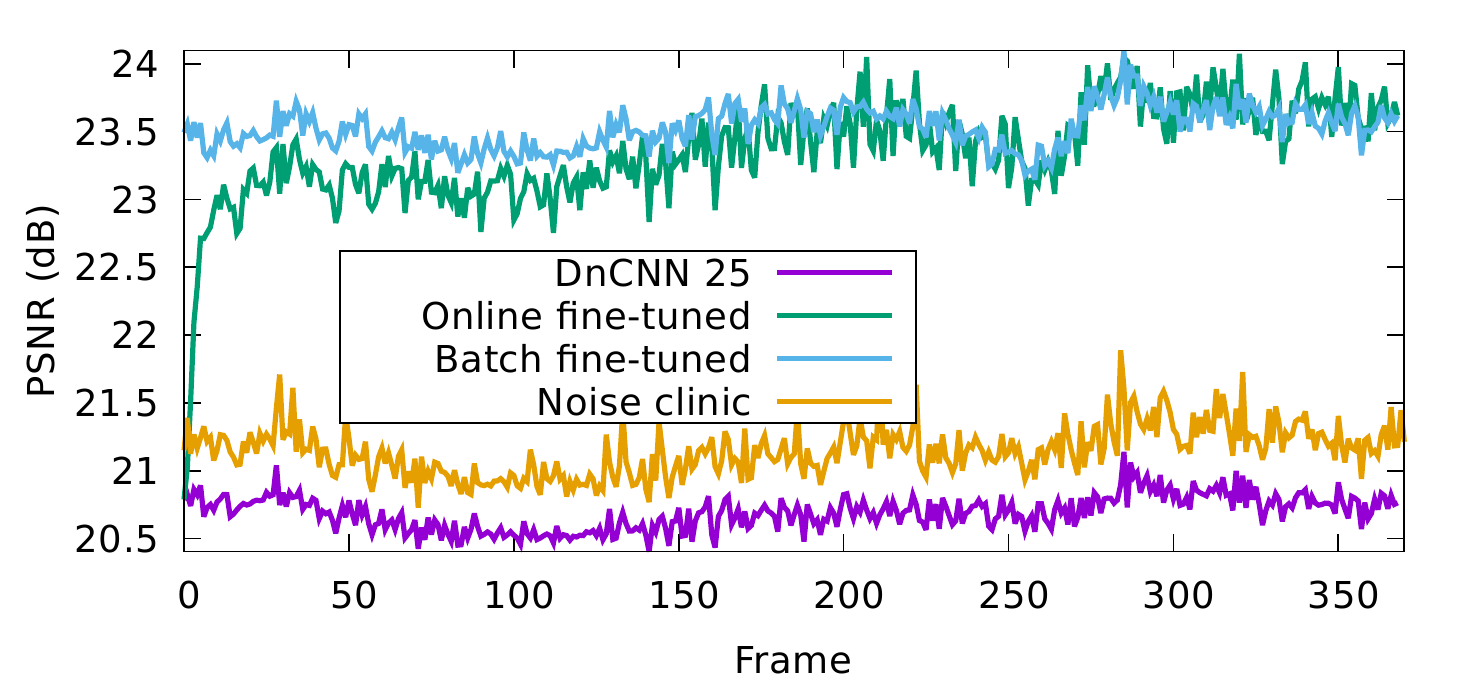}
\includegraphics[width=\columnwidth]{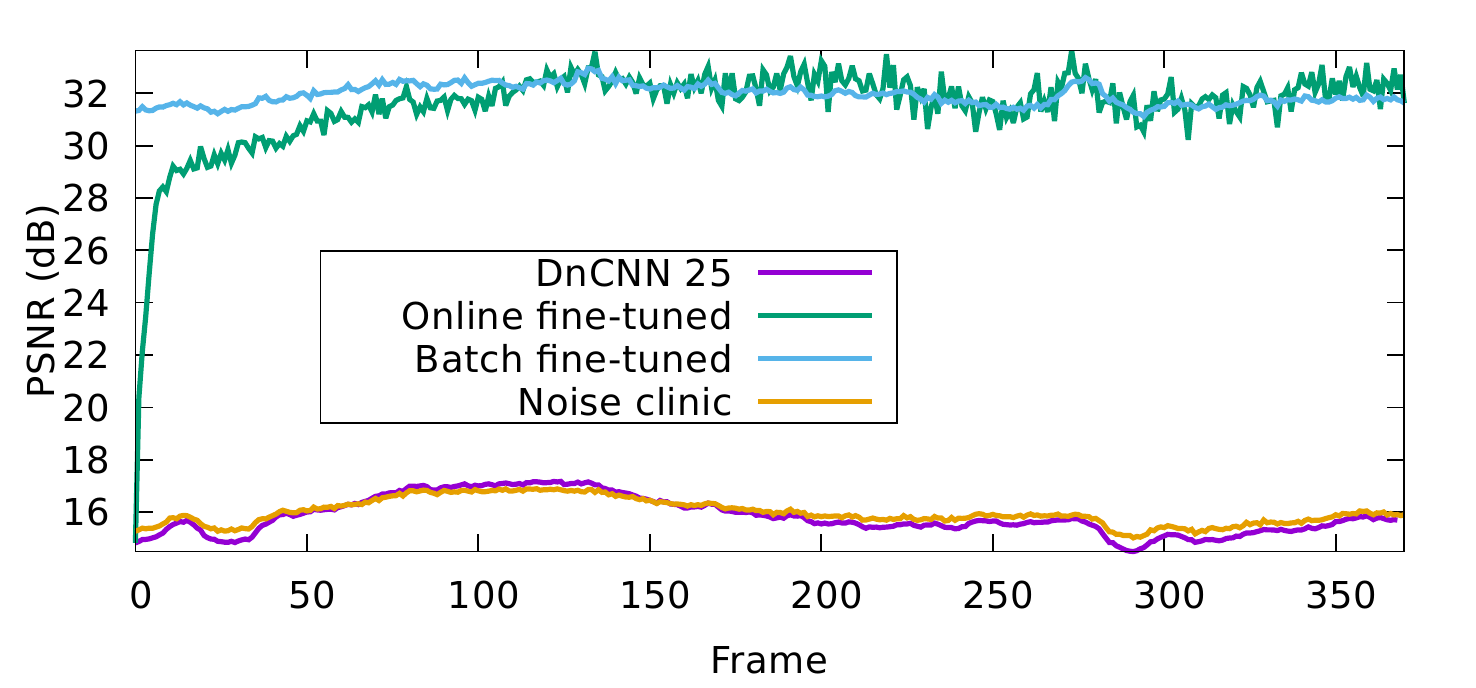}
\includegraphics[width=\columnwidth]{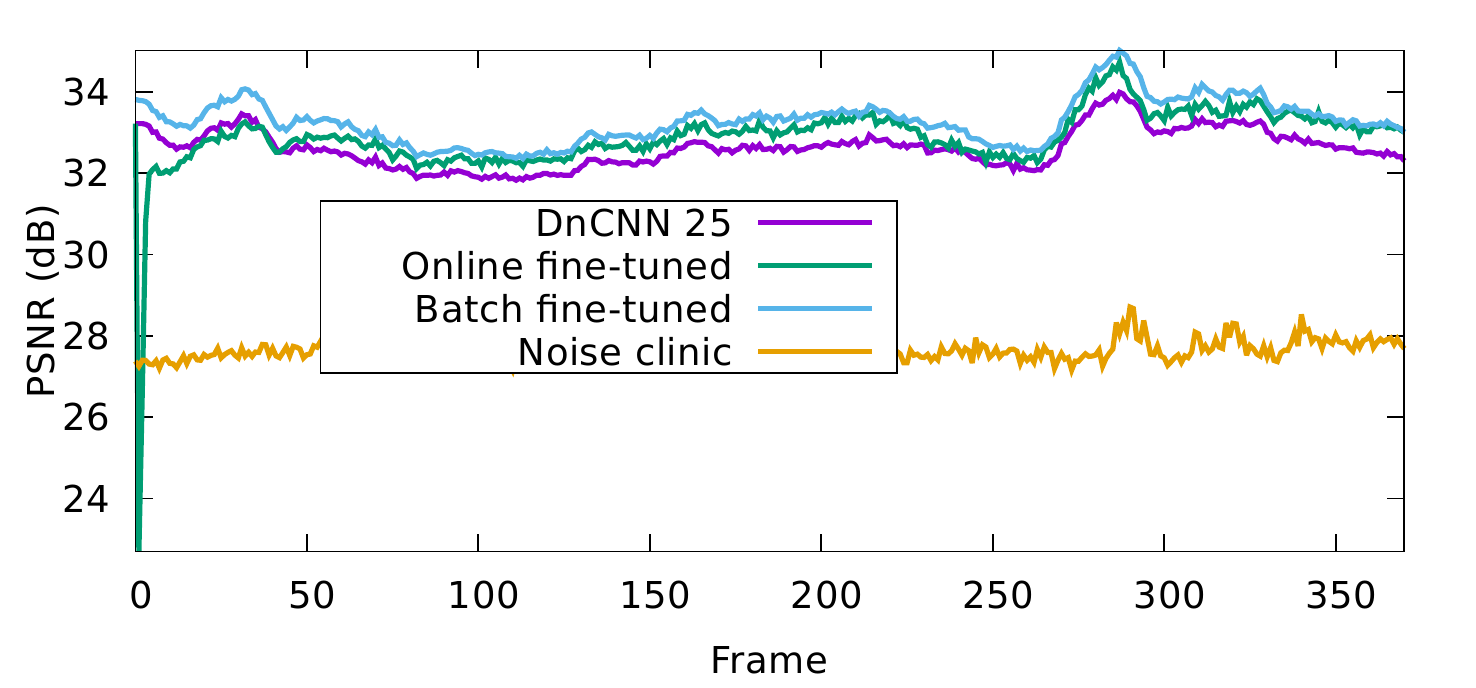}
\caption{Different types of noise. From top-left to bottom right: multiplicative Gaussian noise, correlated Gaussian noise, salt and pepper noise, Gaussian noise after JPEG compression. The fine-tuned network (both online and batch) always performs better than the original network.}
\label{fig:exp_other_noises}
\end{figure*}

We have also tested the proposed fine-tuning on four other types of noise: multiplicative Gaussian, correlated, salt and pepper and compressed Gaussian. We  present the corresponding per-frame PSNRs in Figure \ref{fig:exp_other_noises}. The multiplicative Gaussian noise is given by
\begin{equation}
f_t(x) = u_t(x) + r_t(x)u_t(x),
\end{equation}
where $r_t(x)$ is white Gaussian noise with standard deviation of $75/255$ (the images are within the range [0,1]).
The resulting variance $\sigma_t^2(x)$ depends on the pixel intensity $u_t(x)$. 
The correlated noise is obtained by convolving AWGN with a disk kernel. The resulting standard deviation is $\sigma=25$. The salt and pepper uniform noise is like the one used \cite{lehtinen2018noise2noise}, obtained by replacing with probability $0.25$ the value of a pixel with a value sampled uniformly in $[0,1]$. Finally, the compressed Gaussian noise, results from compressing an image corrupted by an AWGN of $\sigma=25$ with JPEG. The last one is particularly interesting because it is a realistic use case for which the noise model is quite hard to estimate \cite{18-gonzales-denoising-decompression}.
While in this case the noise can be generated synthetically for training a network over a dataset, this is not possible with other compression tools (for example for proprietary technologies). 
We can see the effectiveness of the fine-tuning for all examples. The off-line training is more stable (smaller variance) and gives slightly better results, although the difference is small.

\begin{figure*}
\centering
\includegraphics[trim={15cm 5cm 5cm 2cm},clip,width=0.25\textwidth]{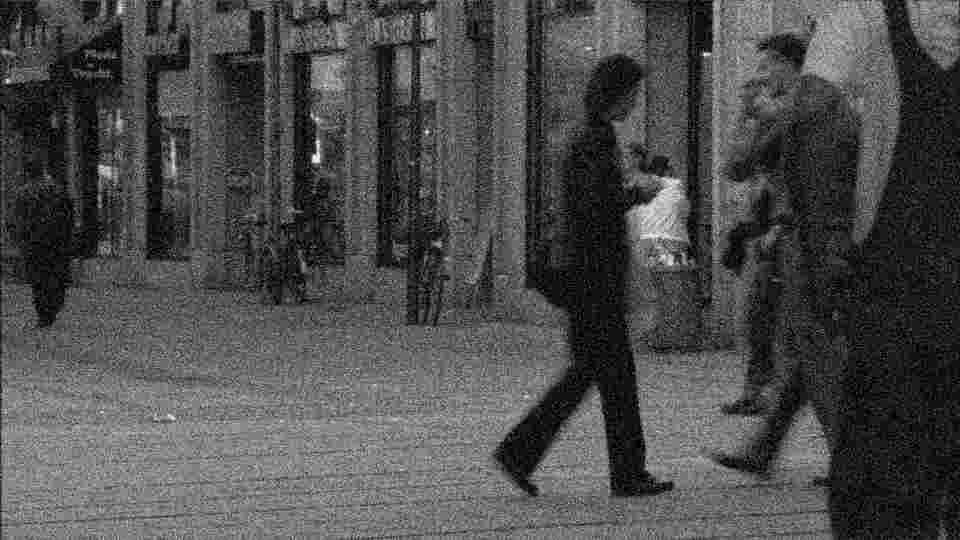}
\includegraphics[trim={15cm 5cm 5cm 2cm},clip,width=0.25\textwidth]{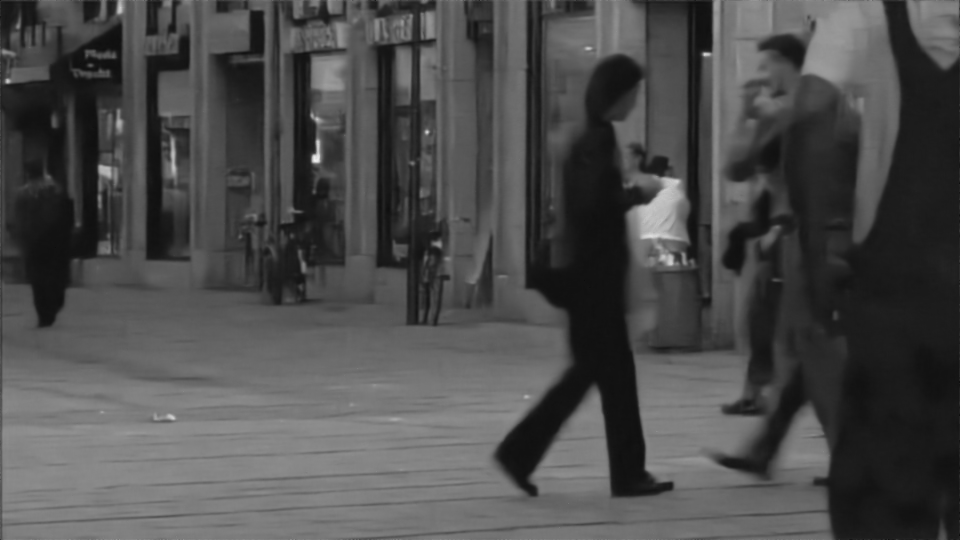}
\includegraphics[trim={15cm 5cm 5cm 2cm},clip,width=0.25\textwidth]{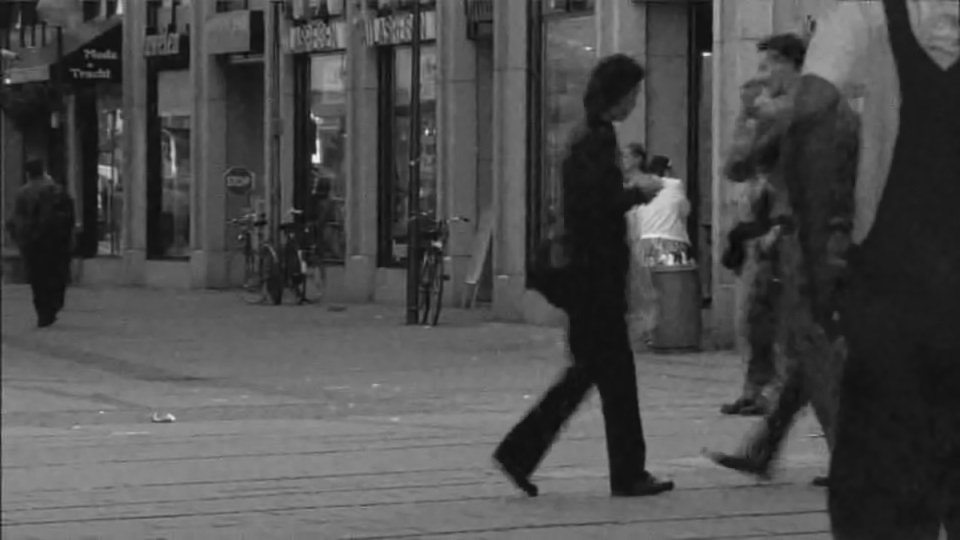}
\includegraphics[trim={15cm 5cm 5cm 2cm},clip,width=0.25\textwidth]{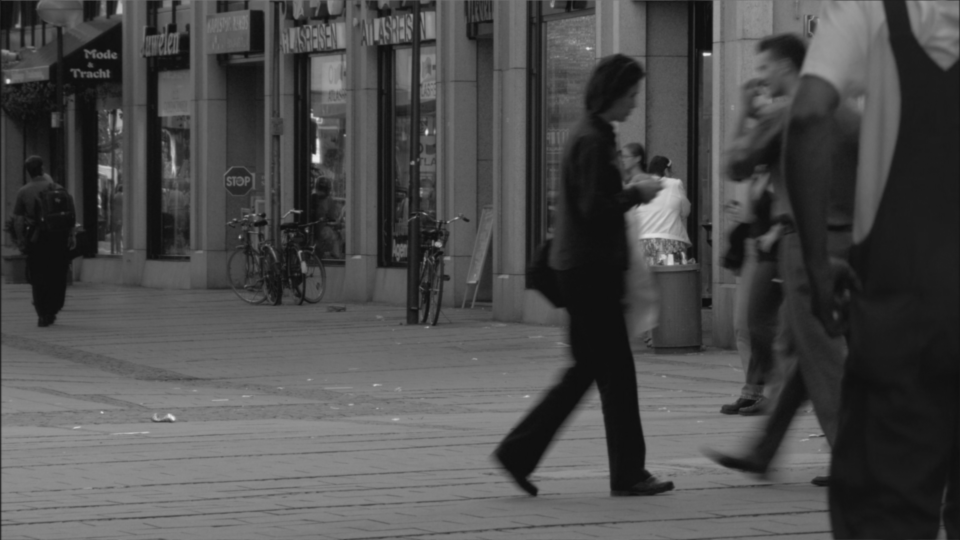}
\includegraphics[trim={15cm 5cm 5cm 2cm},clip,width=0.25\textwidth]{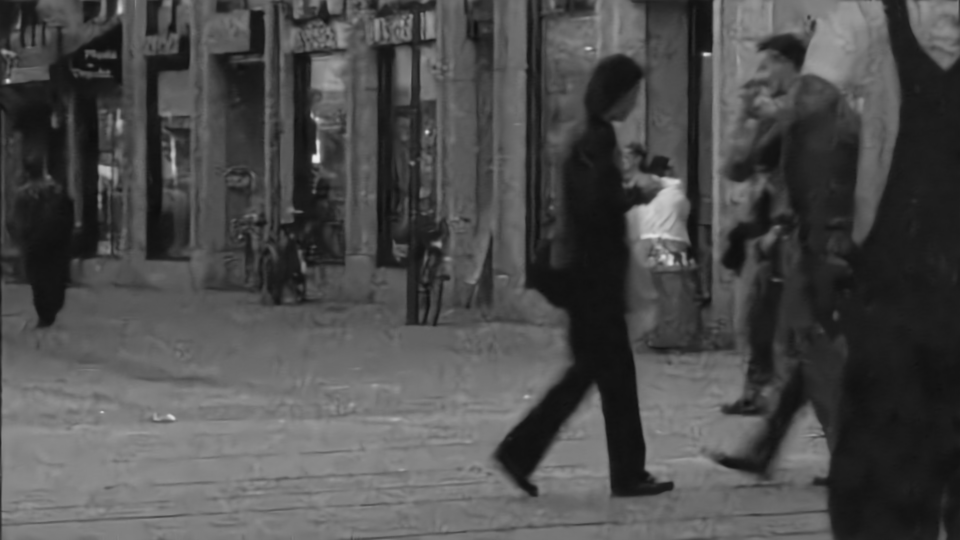}
\includegraphics[trim={15cm 5cm 5cm 2cm},clip,width=0.25\textwidth]{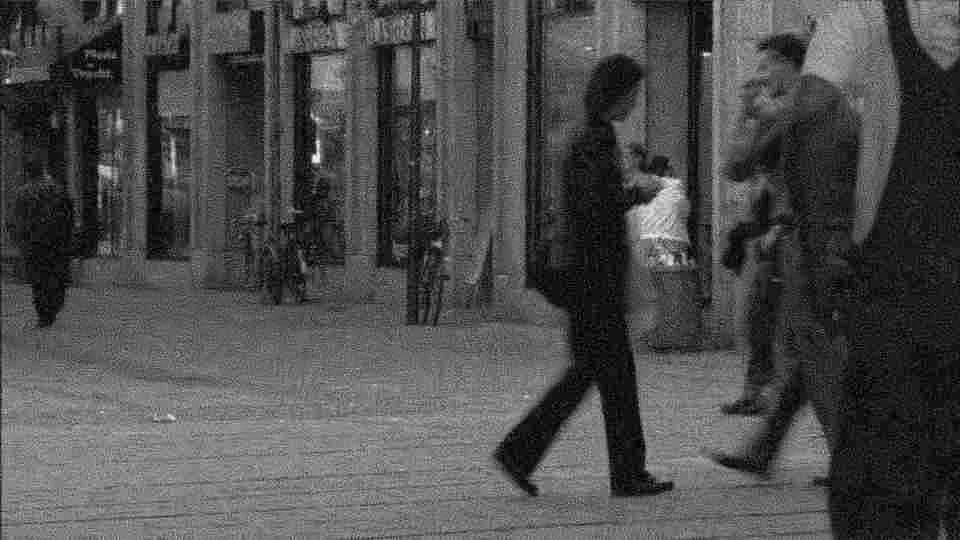}
\caption{Example of denoising of an image corrupted by a JPEG compressed Gaussian noise. The fine-tuned network doesn't produce any visible artifacts, contrary to the original DnCNN used for the fine-tuning process. From left to right, top to bottom: Noisy, fine-tuned, VBM3D, ground truth, DnCNN trained for a Gaussian noise, noise clinic.}
\label{fig:visual_jpeg}
\end{figure*}

A visual comparison with other methods is shown in Figure \ref{fig:visual_jpeg} for JPEG compressed noise and in Figure \ref{fig:visual_g50} for AWGN with $\sigma=50$.
Visual examples on real data are presented in the supplementary material.
The result of the fine-tuned network has no visible artifacts and is visually pleasing even though the network has never seen this type of noise before the fine-tuning. A limitation of the method is the oversmoothing of texture. Indeed DnCNN has a tendency of oversmoothing textures. Using a network designed for video denoising should help recover more texture and improve temporal consistency \cite{vnlnet}. Another cause is the optical flow. Since it is computed on downscaled noisy frames it is imprecise around edges. This leads to false correspondences between frames and introduces some blur. Improved registration should lead to sharper results.

In Tables \ref{tab:gaussian_noise_n50} and \ref{tab:gaussian_jpeg_n25} we show the PSNR of the results obtained on 4 sequences for AWGN of $\sigma=50$ and JPEG compressed AWGN of $\sigma = 25$ and compression factor $10$. For the case of AWGN the fine-tuned networks attain the performance of the DnCNN trained for that specific noise. For JPEG compressed Gaussian noise, the batch fine-tuned network is on average $0.3dB$ above the pre-trained network. 

\begin{figure*}
\centering
\includegraphics[trim={15cm 5cm 5cm 2cm},clip,width=0.24\textwidth]{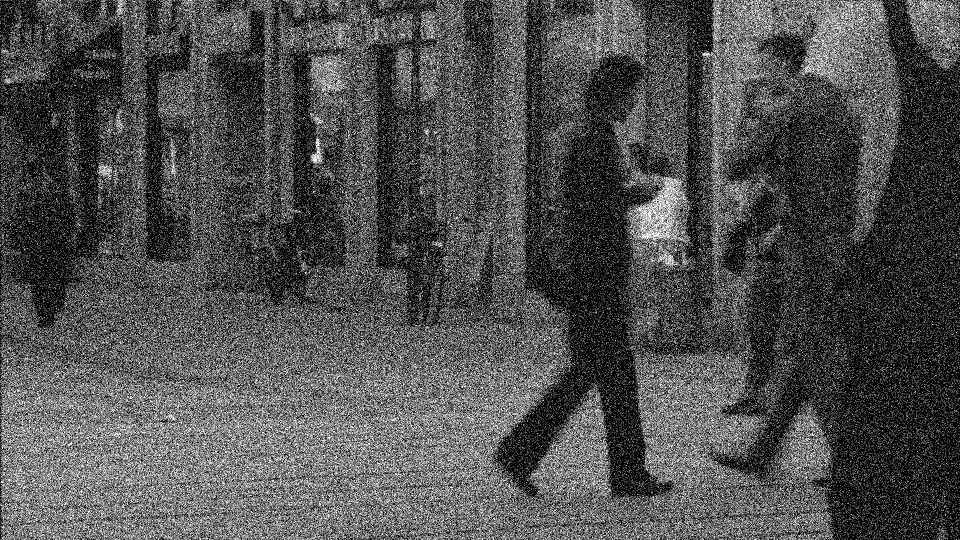}
\includegraphics[trim={15cm 5cm 5cm 2cm},clip,width=0.24\textwidth]{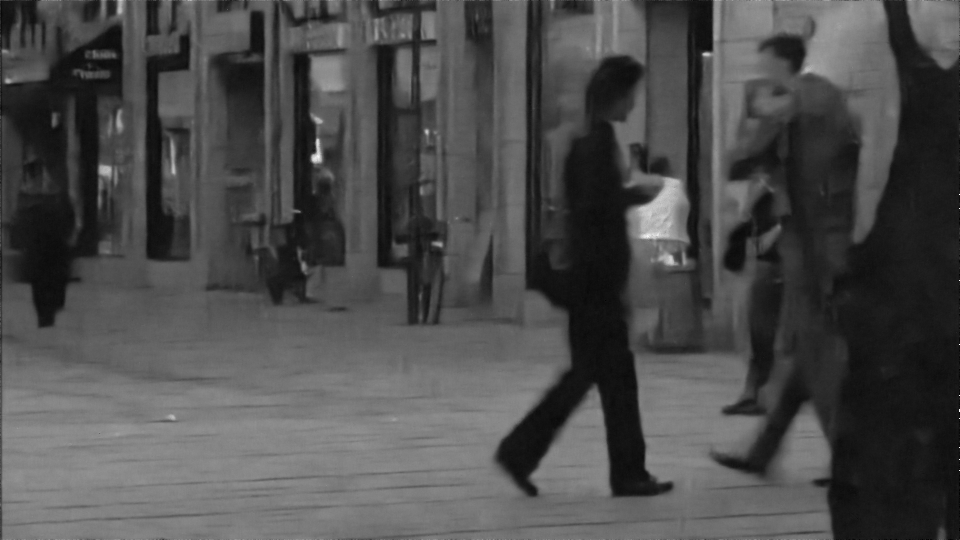}
\includegraphics[trim={15cm 5cm 5cm 2cm},clip,width=0.24\textwidth]{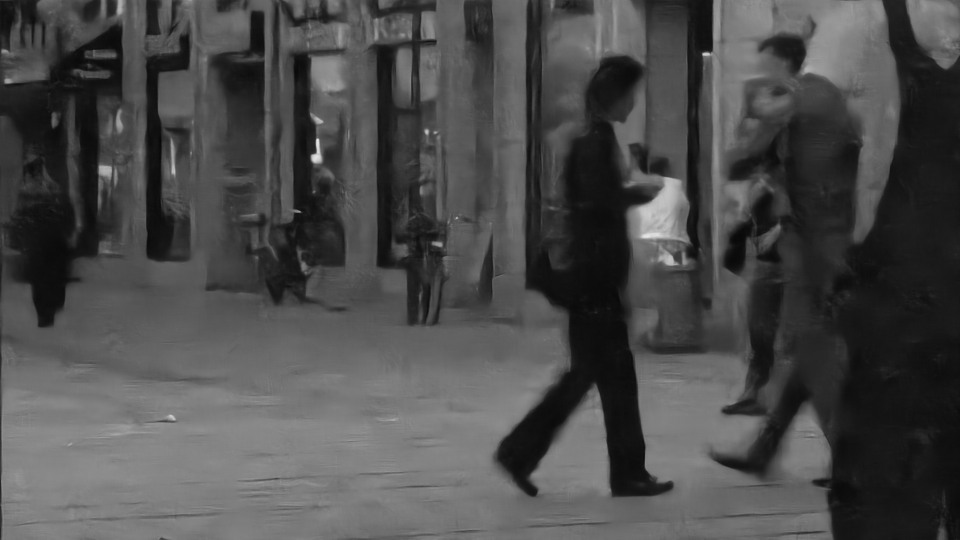}
\includegraphics[trim={15cm 5cm 5cm 2cm},clip,width=0.24\textwidth]{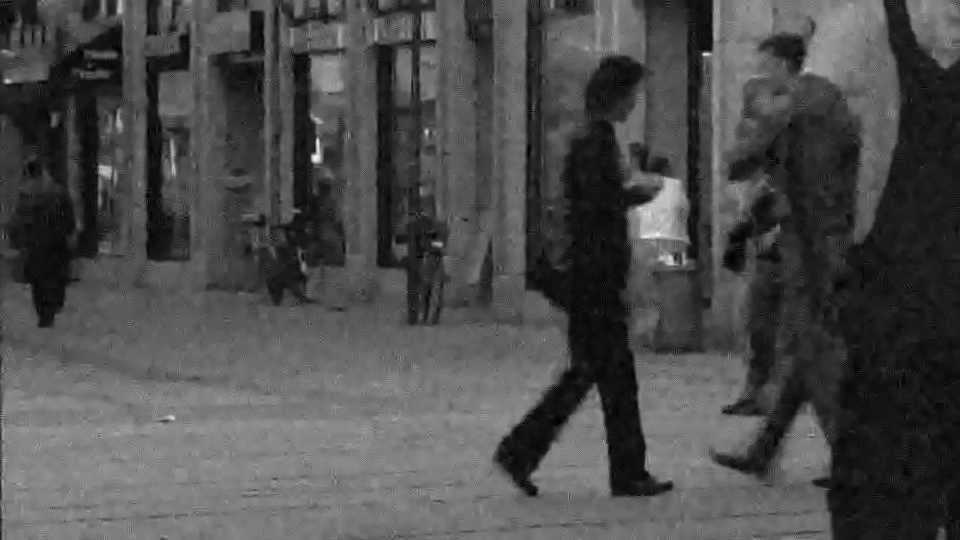}

\includegraphics[trim={15cm 5cm 5cm 2cm},clip,width=0.24\textwidth]{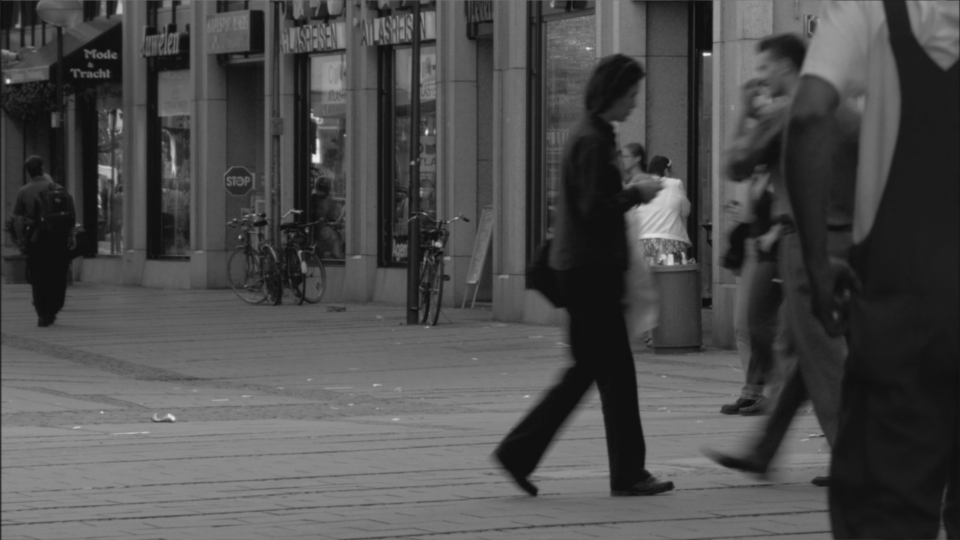}
\includegraphics[trim={15cm 5cm 5cm 2cm},clip,width=0.24\textwidth]{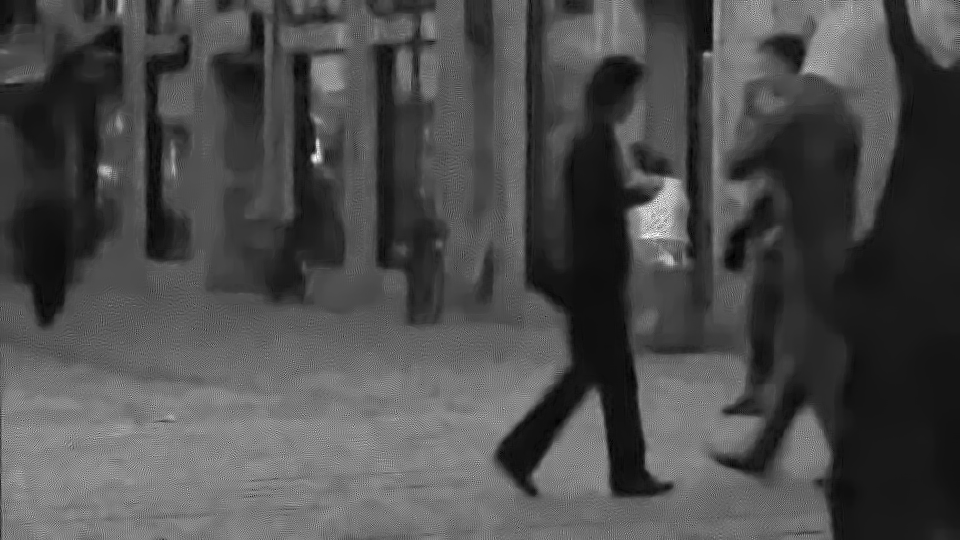}
\includegraphics[trim={15cm 5cm 5cm 2cm},clip,width=0.24\textwidth]{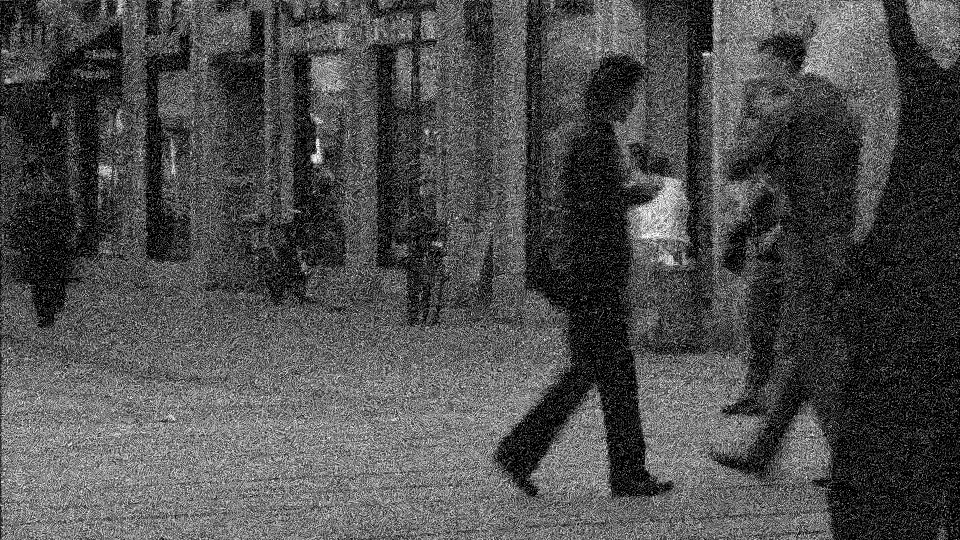}
\caption{Example of denoising of an image corrupted by a Gaussian noise of standard deviation $\sigma=50$. The fine-tuned network doesn't produce any visible artifact, the results are comparable to a DnCNN trained for this particular type of noise. From left to right, top to bottom: Noisy, fine-tuned, DnCNN trained for a Gaussian noise with $\sigma=50$, VBM3D, ground truth, noise clinic, DnCNN trained for a Gaussian noise with $\sigma=25$.}
\label{fig:visual_g50}
\end{figure*}

\begin{table}
	\begin{center}
		{\small
		\renewcommand{\tabcolsep}{1.6mm}
        \renewcommand{\arraystretch}{1.3}
        \begin{tabular}{ @{}l  c  c  c  c @{\hskip 0.7cm} c @{}}
            \toprule
            Method               &     walk  &      crowd    & football &     station   &     Average  \\
            \midrule
            DnCNN 25             &        17.02          &           11.24  &        15.09   &        13.86  &        14.30 \\
            DnCNN 50             &        31.02          &     \Best{25.83} &  \Best{31.67}  &        30.09  &        29.65 \\
            Online fine-tuned    &        30.84          &           25.58  &        31.33   &        29.90  &        29.59 \\
            Batch fine-tuned     &  \Best{31.22}         &     \Best{25.83} &        31.54   &  \Best{30.39} &  \Best{29.75}\\
            Noise Clinic         &        23.85          &           22.13  &        24.57   &        24.39  &        23.74 \\
            \textit{VBM3D}       &\textit{31.57}         &   \textit{27.02} &\textit{31.97}  &\textit{31.33} &\textit{30.47}\\
            \bottomrule
        \end{tabular}}
    \end{center}
    \caption{PSNR values for 4 sequences with AGWN of standard deviation $\sigma=50$.}
	\label{tab:gaussian_noise_n50}
\end{table}

\begin{table}
	\begin{center}
		{\small
		\renewcommand{\tabcolsep}{1.6mm}
        \renewcommand{\arraystretch}{1.3}
        \begin{tabular}{ @{}l  c  c  c  c @{\hskip 0.7cm} c @{}}
            \toprule
             Method               & walk &    crowd  & football &    station    & Average      \\
            \midrule
             DnCNN 25             &        32.62    &  \Best{27.31} &        32.48   &        31.48  &        30.97 \\
             Online fine-tuned    &        32.86    &        27.20  &        32.79   &        30.88  &        30.94 \\
             Batch fine-tuned     &  \Best{33.28}   &        27.19  &  \Best{32.91}  &  \Best{31.58} &  \Best{31.24}\\
             Noise Clinic         &        27.62    &        25.17  &        27.20   &        26.89  &        26.72 \\
             \textit{VBM3D}                &\textit{34.16}   &\textit{28.95} &\textit{33.83}  &\textit{33.53} &\textit{32.62}\\
            \bottomrule
        \end{tabular}}
    \end{center}
    \caption{PSNR values on JPEG compressed AWGN noise with $\sigma = 25$ and compression factor $10$.}
	\label{tab:gaussian_jpeg_n25}
\end{table}

\begin{figure}
\centering
\includegraphics[width=\columnwidth]{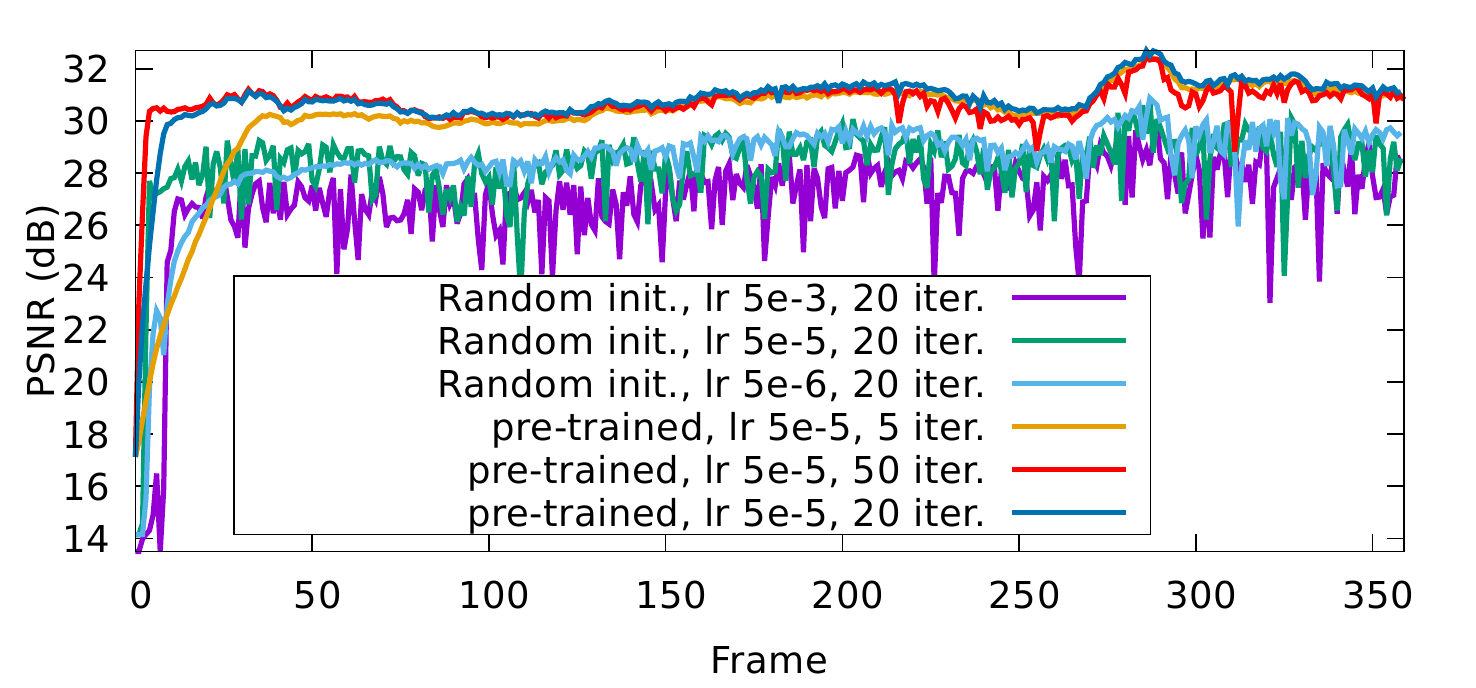}
\includegraphics[width=\columnwidth]{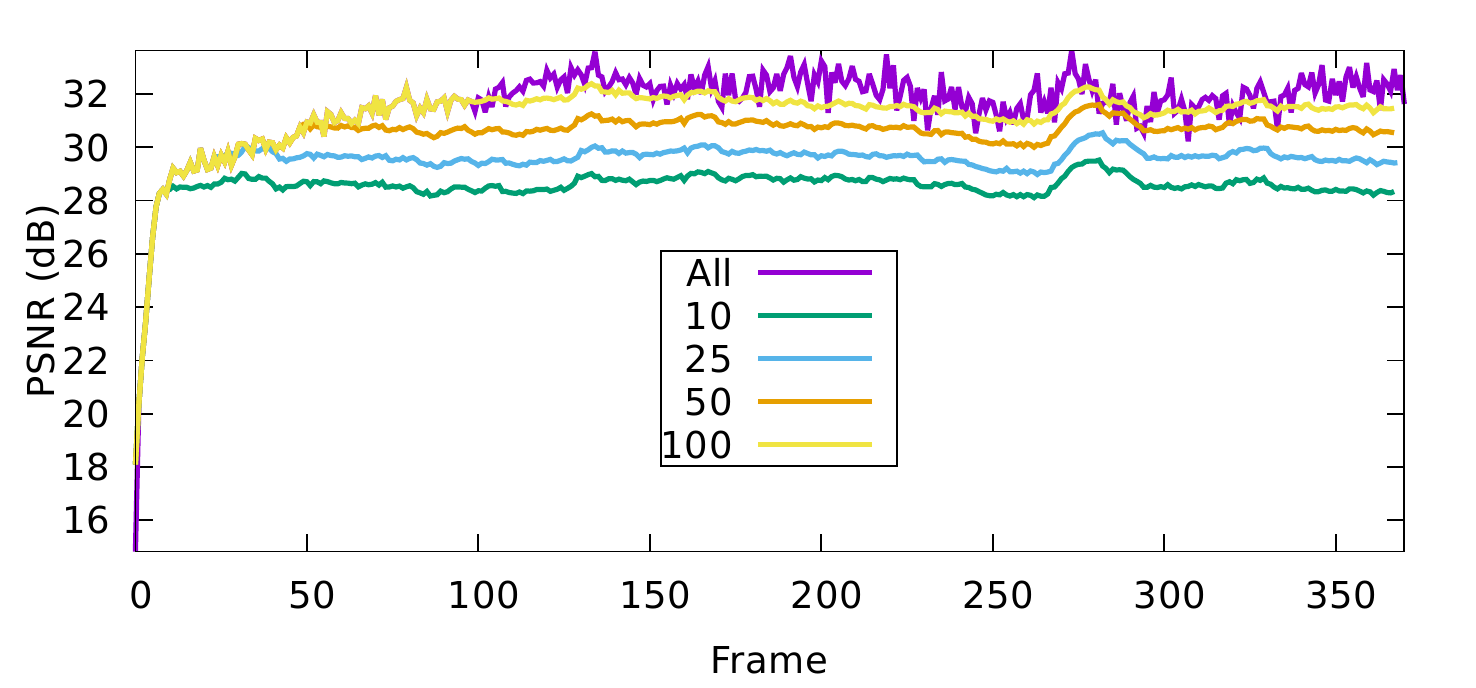}
\caption{Impact of parameters. Top: Impact of the learning rate and the number of iterations. It also shows the gap between using a pre-trained network and random initialization. Bottom: Impact of the number of frames used for fine-tuning.}
\label{fig:exp_parameters}
\end{figure}

Figure \ref{fig:exp_parameters} shows the impact of different parameters of the method.
The main parameters of the proposed training are the learning rate and the number of per-frame iterations. Fewer iterations require more frames for convergence. In turn the result has smaller variance. A similar analysis can be done for the learning rate. We also show the importance of using a pre-trained network compared to a random initialization. There is a $2dB$ gap in favor of the pre-trained network. 
The other important parameter is the number of frames used for fine-tuning. The fine-tuning is stopped at a frame $t_0$ and $\theta^{\text{ft}}_{t_0}$ is used to process the remaining frame. We can see that the more frames used for the fine-tuning the better the performance. 

\begin{figure}
\centering
\includegraphics[width=\columnwidth]{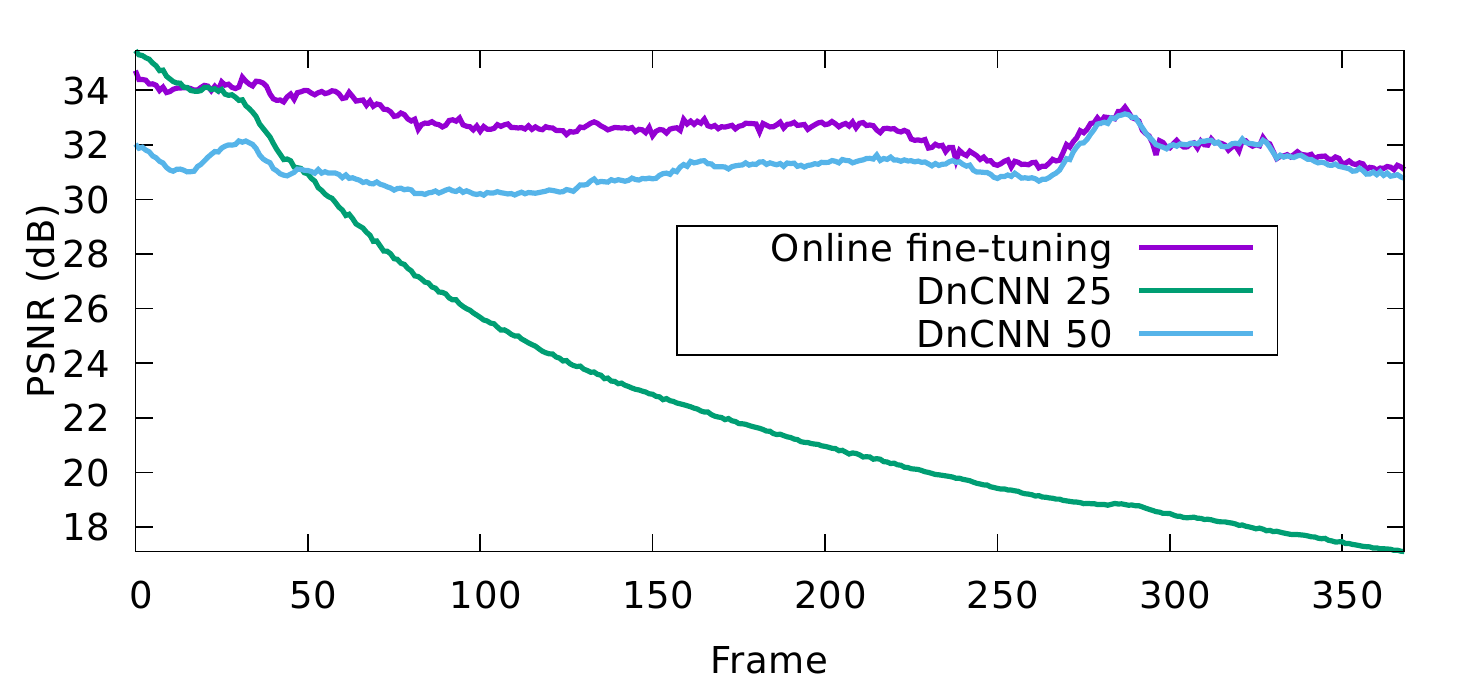}
\includegraphics[width=\columnwidth]{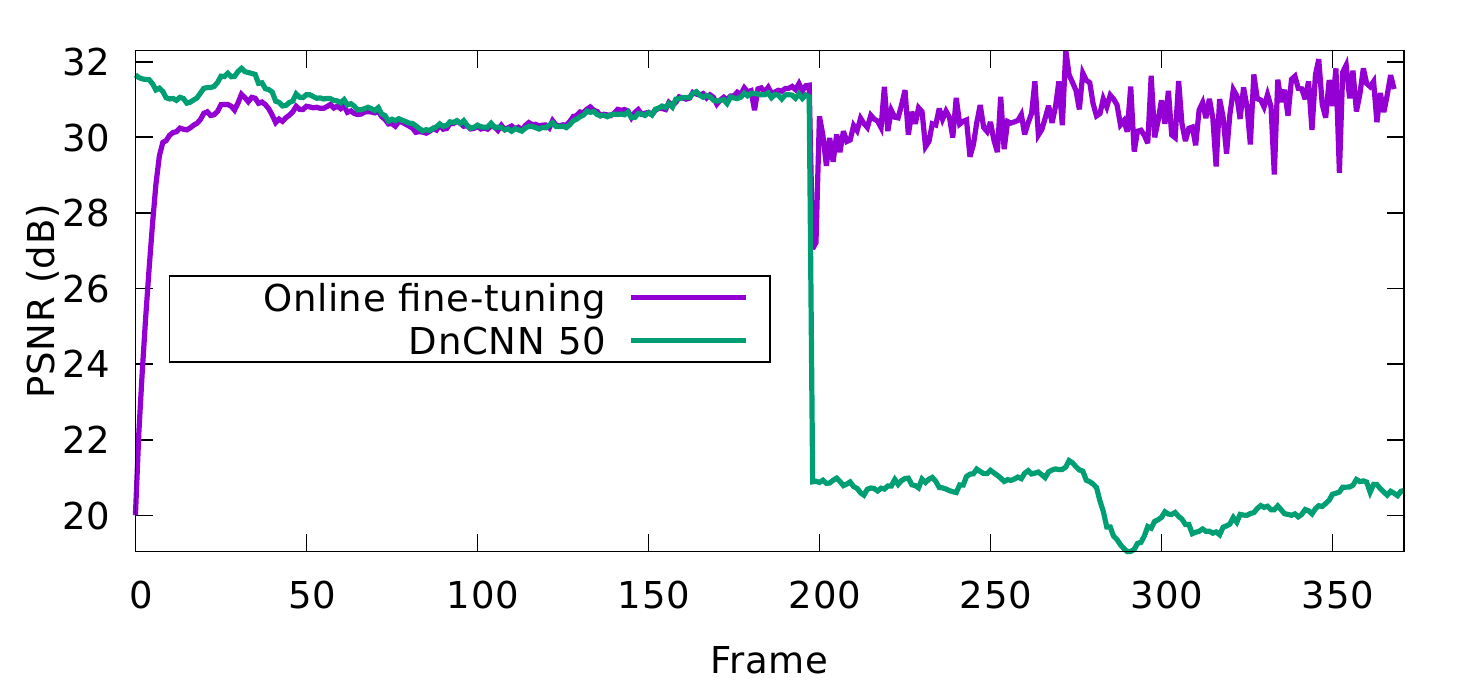}
\caption{Lifelong learning. Top: Slow change. Bottom: sudden change. The fine-tuned network adapts without difficulty to slow changes and sudden changes. See text for more details}
\label{fig:exp_evolving}
\end{figure}

Finally Figure \ref{fig:exp_evolving} shows examples of lifelong learning. The first example shows a slowly evolving noise (starting with a Gaussian noise with standard deviation $\sigma=25$ that linearly increases up to $\sigma=50$). The fine-tuned network performs better than the two reference networks for respectively $\sigma=25$ and $\sigma=50$. The second example shows a sudden change (starting with a Gaussian noise with standard deviation $\sigma=50$ and changes to Salt and pepper noise at frame $200$). In that case the fine-tuned network adapts quickly to the new noise model.

The running time depends on the network.
We used DnCNN but other networks can be used instead and trained with the proposed method. Each fine-tuning iteration runs a back-propagation step which takes $0.33s$ on a NVIDIA Titan XP for DnCNN for a $960\times 540$ frame. Fifty frames with $20$ iterations per frame take $5$ mins. For comparison, training DnCNN from scratch over a dataset requires around $6h$ (and a dataset). By using a lighter network and reducing the per-frame-iterations it might be possible to achieve real time frame rates. Moreover, the fine-tuning can be done on a fixed number of frames at the beginning or run in the background each number of frames for cases when the computational efficiency is important.

\section{Discussion and perspectives}
\label{sec:conclusion}
Denoising methods based on deep learning often require large datasets to achieve state-of-the-art performance. Lehtinen \etal~\cite{lehtinen2018noise2noise} pointed out that in many cases the clean ground truth images are not necessary, thus simplifying the acquisition of the training datasets. With the framework presented in this paper we take a step further and show that a single video is often enough, removing the need for a dataset of images. By applying a simple frame-to-frame training on a generic pre-trained network (for example a DnCNN network trained for  additive Gaussian noise with fixed standard deviation), we successfully denoise a wide range of different noise models even though the network has \emph{never} seen the video nor the noise model before its fine-tuning. This opens the possibility to easily process data from any unknown origin.

We think that the current fine tuning process can still be improved. First, given that the application is video denoising, it is expected that better results will be achieved by a video denoising network (the DnCNN network processes each frame independent of the others). Using the temporal information could improve the denoising quality, just like video denoising methods improve over frame-by-frame image denoising methods, but also might stabilize the variance of the result for the on-line fine-tuning.

\section*{Acknowledgment}

The authors gratefully acknowledge the support of NVIDIA Corporation with the donation of the Titan Xp GPU used for this research. Work partly financed by IDEX Paris-Saclay IDI 2016, ANR-11-IDEX-0003-02, Office of Naval research grant N00014-17-1-2552, DGA Astrid project \guillemotleft filmer la Terre\guillemotright~n$^o$ ANR-17-ASTR-0013-01, MENRT.

{\small
\bibliographystyle{ieee}
\bibliography{main}
}

\end{document}